# Digital Twin approach to Clinical DSS with Explainable AI

Reducing subjectivity in liver disease diagnosis using Domain Knowledge and Machine Learning.

Dattaraj Jagdish Rao
Persistent Systems Ltd.
dattaraj_rao@persistent.co.in

Shraddha Mane
Persistent Systems Ltd.
shraddha_mane@persistent.co.in

**ABSTRACT**

*We propose a digital twin approach to improve healthcare decision support systems with a combination of domain knowledge and data. Domain knowledge helps build decision thresholds that doctors can use to determine a risk or recommend a treatment or test based on the specific patient condition. However, these assessments tend to be highly subjective and differ from doctor to doctor and from patient to patient. We propose a system where we collate this subjective risk by compiling data from different doctors treating different patients and build a machine learning model that learns from this knowledge. Then using state-of-the-art explainability concepts we derive explanations from this model. These explanations give us a summary of different doctor domain knowledge applied in different cases to give a more generic perspective. Also these explanations are specific to a particular patient and are customized for their condition. This is a form of a digital twin for the patient that can now be used to enhance decision boundaries for earlier defined decision tables that help in diagnosis. We will show an example of running this analysis for a liver disease risk diagnosis. All the code used in this paper is available at the link below on Google Colab as a notebook.*

*https://colab.research.google.com/drive/1Q3Eop0e26_S2QueGHL_jalGJs3T5CR_O*

Keywords: explainable ai, decision support system, machine learning, digital twin

**INTRODUCTION**

Clinical Decision Support Systems (CDSS) provides clinicians, staff and patients with knowledge and person-specific information at the point of care. The focus is to bring insights, best-practices and recommendations from historical data from electronic health records (EHR) to help the care-provider take informed decisions and improve the quality of care while reducing costs. The insights need to be highly customized to a particular individual being treated in order to take into consideration factors like side-effects, allergies, response to treatments, etc. and avoid time on ineffective tests and procedures for that patient. Traditionally these CDSS systems have been a way to document the knowledge in the brains of care-providers and capture it inside a Software system. This has been effective however many empirical decisions made remain undocumented since they are not captured formally as rules. Also, the decisions are often subjective between doctor to doctor. If we can eliminate the subjectivity through data science and devise a generic system that integrates the logic of several individual decisions by doctors for specific patient cases. Also, this system should be transparent to inform the doctor why a particular decision was made. This is what we try and define by a Digital Twin. Here is where a data science approach is needed to extract these valuable patient-specific insights from historical data and have these working alongside deterministic rules to provide the best care suggestions for a particular individual.

A Digital Twin can be defined as a dynamic digital replica of the patient, created with data that is historically available. It is also designed to capture data continuously from the life of that individual. A digital twin is intended for more effective care interventions by helping clinicians and other intersecting care technologies to really "know" the patient. A digital twin data architecture dives deep to help characterize the patient's uniqueness, such as: medical condition, response to drugs, therapy, ecosystem, etc. Digital Twins try to build highly customized models specific to a particular patient – defined by deterministic rules as well as patterns learned from data using Machine Learning (ML) models. This way the Twins try to take a best of both worlds approach leveraging clinical knowledge and patterns seen from historical data – to empower care providers with all relevant data and evidence to make the right decision for the patient.

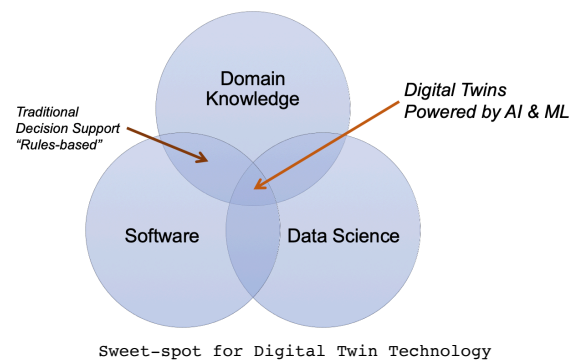

Sweet-spot for Digital Twin Technology

The major difference behind traditional ML models and digital twins is the customization for a particular patient. The models developed are patient-specific and need to be continuously updated with the latest patient data to be effective. They can then be used by care providers to run analysis and what-if scenarios to understand how the patient will react to different drugs and procedures. The virtual Software-defined twin can provide valuable insights to physicians to plan the right care plans for the patients – backed with historical evidence and latest data.

As seen from figure above there are 3 major aspects of building a patient-specific Digital Twin:

1. Domain Knowledge
2. Data Science
3. Software that integrates domain with data

Let's look at each of this in detail with a simple example of determining risk of Liver disease by observing certain test results.

## DOMAIN KNOWLEDGE

Domain Knowledge of how to diagnose a condition, analyze symptoms and recommend appropriate tests and procedures is the key ingredient of a CDSS. This knowledge may be recorded into medical documents and then encoded into Software as algorithms and rules. Modeling techniques like Decision Modeling and Notation (DMN) helps capture this knowledge in a structured form. Below figures shows a decision modeled in Camunda BPMN using DMN notation for deciding on the risk for liver disease based on patients age and 2 test results for Alanine transaminase (ALT) and Aspartate transaminase (AST). This is a pretty simple case, but as rules get complex, it is very difficult for care providers to memorize them all. Hence written manuals with these rules or CDSS which can automatically provide such recommendations is extremely helpful. A Digital Twin system should start with this knowledge and try to capture it a readable and verifiable system and incorporate into Software.

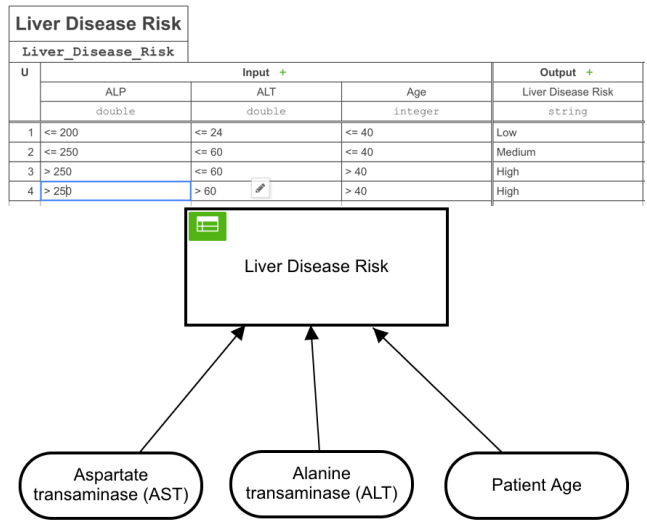

## DATA SCIENCE

While basic rules can be captured and documented by reading medical documents and interviewing doctors, it becomes very difficult to form fixed deterministic rules as we get more and diverse datasets. Here we resort to data-driven techniques like Machine Learning (ML) to extract patterns from data. Below is a dataset of liver tests several patients and an indication of Risk marked by consultation with human care provider. This dataset is the Indian Liver Patient Dataset made openly available by the UC Irvine Machine Learning Repository [1][2].

|   | Age | Gender | Total Bilirubin | Direct Bilirubin | ALP | ALT | AST | Total Protiens | Albumin | Albumin/Globulin Ratio | Risk |
|---|-----|--------|-----------------|------------------|-----|-----|-----|----------------|---------|-----------------------|------|
| 0 | 65  | 0      | 0.7             | 0.1              | 187 | 16  | 18  | 6.8            | 3.3     | 0.90                  | 0    |
| 1 | 62  | 1      | 10.9            | 5.5              | 699 | 64  | 100 | 7.5            | 3.2     | 0.74                  | 0    |
| 2 | 62  | 1      | 7.3             | 4.1              | 490 | 60  | 68  | 7.0            | 3.3     | 0.89                  | 0    |
| 3 | 58  | 1      | 1.0             | 0.4              | 182 | 14  | 20  | 6.8            | 3.4     | 1.00                  | 0    |
| 4 | 72  | 1      | 3.9             | 2.0              | 195 | 27  | 59  | 7.3            | 2.4     | 0.40                  | 0    |

Dataset of historical patient data with Risk marked by human

We can now apply ML classification algorithms like Logistic Regression, Neural Networks, Support Vector Machines, etc. to build a classifier that takes the patient data as input and predicts risk. Here we are trying to collectively build a model that can learn all the latent patterns in data including combined thought-process of care-provider in determining the overall risk. We trained a random forest classification model on the data that learned to predict a patient as risk or no-risk of liver disease. Below is the learning curve for classifier with an accuracy of 0.72 on validation dataset. We had limited data of 583 samples of which we did a 80-20 split. With more data the model will generalize with much higher accuracy.

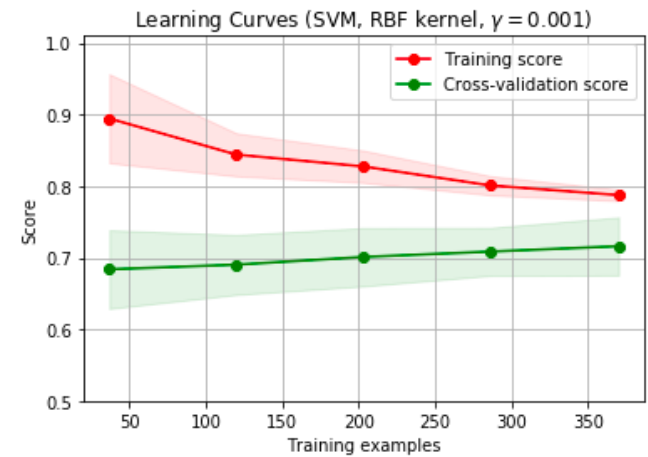

The problem with ML model like this one is that it will act like a black-box, giving you a level of risk based on the patient data entered. It will not tell you what factors will be considered while making such a decision.

In critical areas like Healthcare, we cannot afford to have black-boxes but need to know what the basis was for making certain decisions. Here is where a relatively new concept called Explainable AI comes in to play. Explainable AI tries to provide not only a decision from the dataset but also a set of supporting evidence to help humans analyze the decision. The final authority to approve or reject the decision is still with the human care-provider.

We will take the random forest (RDF) model we developed and use a state-of-the-art Explainable AI library called LIME (Local Interpretable Model-Agnostic Explanations) [3] to try and generate explanations.

Below are 2 examples of explanations generated by LIME. We see that it analyzes the input data, makes a risk prediction and highlights the factors that affected this risk score.

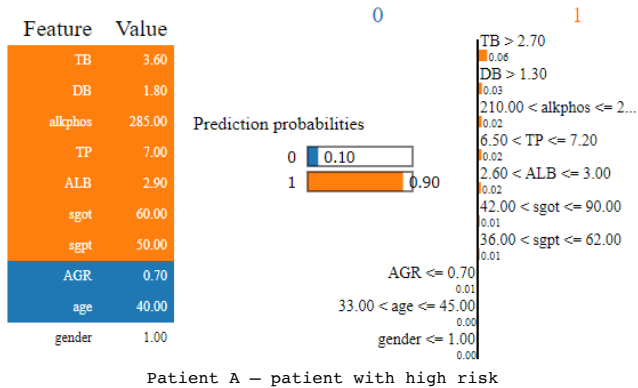

Patient A – patient with high risk

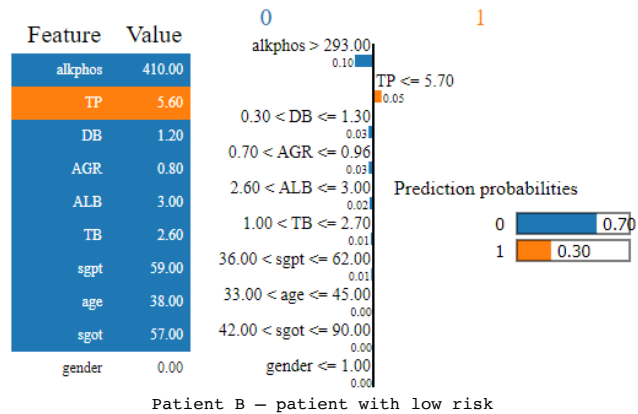

Patient B – patient with low risk

The explanations shown above show the contribution of top features to the prediction and are specific to each patient data record, as we saw earlier from our definition of Digital Twins. Also, it highlights which features are contributing towards positive decision, which towards negative decision and which don't contribute in decision making at all. For e.g. we can see from above explanations that gender value does not affect the risk decision made by the model.

We also used Partial Dependence Plot (PDP) method to find relation between feature and model's output. It shows effect of feature on the prediction made by machine learning model. Below are two features – Sgpt Alamine Aminotransferase and gender PDP plot. First, we fit machine learning model and then analyze the partial dependencies. In this case we have fitted logistic regression model to predict risk of liver disease and use partial dependence plot to understand the relationships that model has learned. The effect of sgpt and gender on the risk of liver disease is as shown in below PDP plots.

For sgpt, PDP shows that in 0 to 130 probability is increasing very rapidly but after 130 it is slowly increasing. Which means if the value of sgpt increases from 0 to 130 then then the risk of liver disease increases by the probability of 0.18. Similarly, if we see PDP of gender then we see there is no significant increment or decrement in the probability by change in a gender value. Thus, there is no relationship between risk prediction model output and gender.

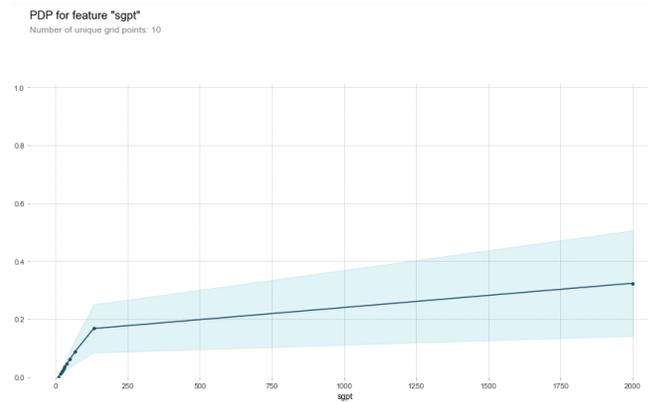

PDP for liver disease risk prediction model and sgpt

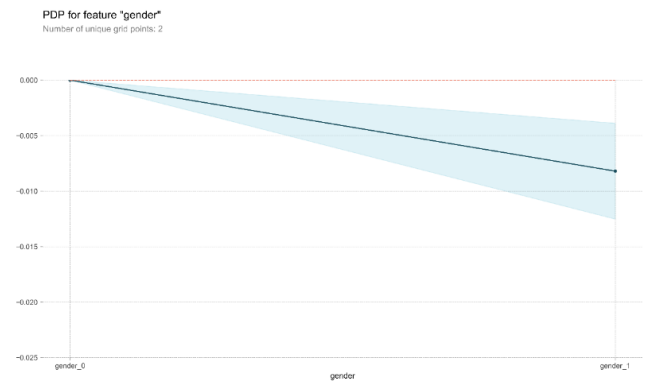

PDP for liver disease risk prediction model and gender

We can now combine these explanations with our knowledge of rules from the DMN system to provide data-backed recommendations for care based on patient test results.

## SOFTWARE

The final piece of the puzzle in the Digital Twin equation is Software. Our Software systems should be able to act like a glue between the domain-based DMN system and the data-based ML system to provide the end user with right recommendations. In simple situations like the one above it's pretty straight forward to extract explanations and match with human decision rules to verify them. However, as systems get very complex and it's difficult to devise human-defined rules, the challenge is to extract rules from data and augment or define new rules.

Key aspect to Digital Twin is that it has to be patient-specific. Hence as new patient data is acquired and processed, the rules need to be adjusted and a customized recommendation is provided. For example, a static human-defined empirical rule may say that patients with ALP less than 200 will show less risk. As we process multiple patient data, if we see that the explanations tend to show that this limiting value for ALP should not be 200 but 175 – then the rule needs to be tuned accordingly. Our Software systems should be able to do this dynamically.

Finally, the key aspect to having a effective Twin is to keep it "alive" with continuous access to real patient data. Our Software systems with the help of data entry screens and APIs should enable this in order to get the best care recommendation.

## CONCLUSION

We saw how clinical DSS systems can be enhanced using concept of Digital Twin that builds a customized per-patient model for providing recommendations. There are 3 key aspects of a Digital Twin. First is domain knowledge which is captured in form of rules and can be modeled using techniques like DMN. With more patient data, we can build ML models to understand patterns and use predictions to help in recommendations. Predictions cannot be black box – we need explainability of models to provide recommendations and evidence on why those were made. This evidence can be tied back to manual rules and can be used to tune them. Finally, Software is the glue that puts these Digital Twins together and keeps them "alive" with real-time data feeds. All these 3 elements together constitute the healthcare digital twin system. We can combine the domain knowledge with multiple types of data like clinical records, sensor readings, social activity, environment factors, etc. to get a true picture of the patient health and recommend the right care at the right time.

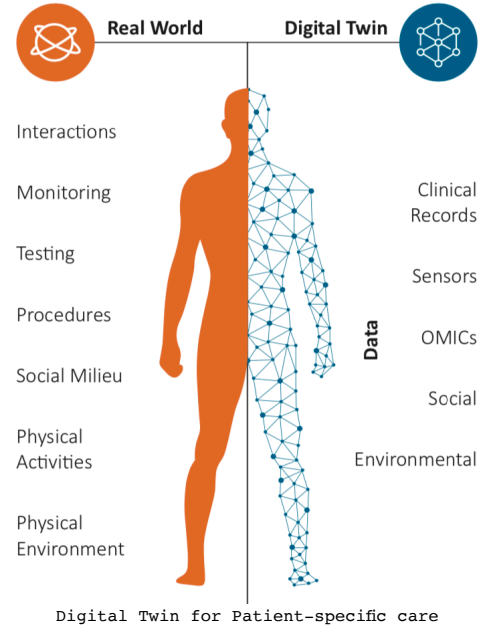

Digital Twin for Patient-specific care